\documentclass[11pt,a4paper]{article}
\usepackage[hyperref]{acl2018}
\usepackage{times}
\usepackage{latexsym}

\usepackage{url}

\aclfinalcopy 
\def\aclpaperid{1520} 

\usepackage{amsfonts,amsmath,amssymb,amsthm,graphicx}
\usepackage[linesnumbered,ruled,vlined]{algorithm2e}
\usepackage{booktabs,multicol,multirow,threeparttable}
\DeclareMathOperator*{\argmin}{arg\,min}
\usepackage{color}
\usepackage[hang,flushmargin]{footmisc}

\title{A La Carte Embedding:\\Cheap but Effective Induction of Semantic Feature Vectors\vspace{0.5cm}}

\author{
	Mikhail Khodak$^\ast$, Nikunj Saunshi$^\ast$ \\
	Princeton University \\
	\texttt{\{mkhodak,nsaunshi\}@princeton.edu} \\
	\And
	\qquad Yingyu Liang \\
	\qquad University of Wisconsin-Madison \\
	\qquad \texttt{yliang@cs.wisc.edu} \\
	\AND
	Tengyu Ma \\
	Facebook AI Research \\
	\texttt{tengyuma@stanford.edu}
	\\\And
	\qquad Brandon Stewart, Sanjeev Arora \\
	\qquad Princeton University \\
	\qquad \texttt{\{bms4,arora\}@princeton.edu}
}

\date{}

\begin{document}
\maketitle

\begin{abstract}
	Motivations like domain adaptation, transfer learning, and feature learning have fueled interest in inducing embeddings for rare or unseen words, $n$-grams, synsets, and other textual features.
	This paper introduces {\em\` a la carte} embedding, a simple and general alternative to the usual word2vec-based approaches for building such representations that is based upon recent theoretical results for GloVe-like embeddings.
	Our method relies mainly on a linear transformation that is efficiently learnable using pretrained word vectors and linear regression.
	This transform is applicable ``on the fly" in the future when a new text feature or rare word is encountered, even if only a single usage example is available. 
	We introduce a new dataset showing how the {\em\` a la carte} method requires fewer examples of words in context to learn high-quality embeddings and we obtain state-of-the-art results on a nonce task and some unsupervised document classification tasks.
\end{abstract}


\section{Introduction}\label{sec:intro}

Distributional word embeddings, which represent the ``meaning" of a word via a low-dimensional vector, have been widely applied by many natural language processing (NLP) pipelines and algorithms \cite{Goldberg:16}.
Following the success of recent neural \cite{Mikolov:13} and matrix-factorization \cite{Pennington:14} methods, researchers have sought to extend the approach to other text features, from subword elements to $n$-grams to sentences \cite{Bojanowski:16,Poliak:17,Kiros:15}.
However, the performance of both word embeddings and their extensions is known to degrade in small corpus settings \cite{Adams:17} or when embedding sparse, low-frequency features \cite{Lazaridou:17}.
Attempts to address these issues often involve task-specific approaches \cite{Rothe:15,Iacobacci:15,Pagliardini:18} or extensively tuning existing architectures such as skip-gram \cite{Poliak:17,Herbelot:17}.

For computational efficiency it is desirable that methods be able to induce embeddings for only those features (e.g. bigrams or synsets) needed by the downstream task, rather than having to pay a computational {\em prix fixe} to learn embeddings for all features occurring frequently-enough in a corpus. 
We propose an alternative, novel solution via {\em\`a la carte} embedding, a method which bootstraps existing high-quality word vectors to learn a feature representation in the same semantic space via a linear transformation of the average word embeddings in the feature's available contexts.
This can be seen as a shallow extension of the distributional hypothesis \cite{Harris:54}, ``a feature is characterized by the words in its context," rather than the computationally more-expensive ``a feature is characterized by the features in its context" that has been used implicitly by past work \cite{Rothe:15,Logeswaran:18}. 

Despite its elementary formulation, we demonstrate that the {\em\`a la carte} method can learn faithful word embeddings from single examples and feature vectors improving performance on important downstream tasks.
Furthermore, the approach is resource-efficient, needing only pretrained embeddings of common words and the text corpus used to train them, and easy to implement and compute via vector addition and linear regression.
After motivating and specifying the method, we illustrate these benefits through several applications:
\begin{itemize}
	\item {\bf Embeddings of rare words:} we introduce a dataset\footnote{Dataset: \url{nlp.cs.princeton.edu/CRW}} for few-shot learning of word vectors and achieve state-of-the-art results on the task of representing unseen words using only the definition \cite{Herbelot:17}.
	\item {\bf Synset embeddings:} we show how the method can be applied to learn more fine-grained lexico-semantic representations and give evidence of its usefulness for standard word-sense disambiguation tasks \cite{Navigli:13,Moro:15}.
	\item {\bf $n$-gram embeddings:} we build seven million $n$-gram embeddings from large text corpora and use them to construct document embeddings that are competitive with unsupervised deep learning approaches when evaluated on linear text classification.
\end{itemize}
Our experimental results\footnote{Code: \url{www.github.com/NLPrinceton/ALaCarte}} clearly demonstrate the advantages of {\em\` a la carte} embedding.
For word embeddings, the approach is an easy way to get a good vector for a new word from its definition or a few examples in context.
For feature embeddings, the method can embed anything that does not need labeling (such as a bigram) or occurs in an annotated corpus (such as a word-sense).
Our document embeddings, constructed directly using {\em\` a la carte} $n$-gram vectors, compete well with recent deep neural representations; this provides further evidence that simple methods can outperform modern deep learning on many NLP benchmarks \cite{Arora:17,Mu:18,Arora:18a,Arora:18b,Pagliardini:18}.


\section{Related Work}\label{sec:related}

Many methods have been proposed for extending word embeddings to semantic feature vectors, with the aim of using them as interpretable and structure-aware building blocks of NLP pipelines \cite{Kiros:15,Yamada:16}.
Many exploit the structure and resources available for specific feature types, such as methods for sense, synsets, and lexemes \cite{Rothe:15,Iacobacci:15} that make heavy use of the graph structure of the Princeton WordNet (PWN) and similar resources \cite{Fellbaum:98}.
By contrast, our work is more general, with incorporation of structure left as an open problem.
Embeddings of $n$-grams are of special interest because they do not need annotation or expert knowledge and can often be effective on downstream tasks.
Their computation has been studied both explicitly \cite{Yin:14,Poliak:17} and as an implicit part of models for document embeddings \cite{Hill:16,Pagliardini:18}, which we use for comparison.
Supervised and multi-task learning of text embeddings has also been attempted \cite{Wang:17,Wu:17}.

A main motivation of our work is to learn good embeddings, of both words and features, from only one or a few examples.
Efforts in this area can in many cases be split into contextual approaches \cite{Lazaridou:17,Herbelot:17} and morphological methods \cite{Luong:13,Bojanowski:16,Pado:16}.
The current paper provides a more effective formulation for context-based embeddings, which are often simpler to implement, can improve with more context information, and do not require morphological annotation.
Subword approaches, on the other hand, are often more compositional and flexible, and we leave the extension of our method to handle subword information to future work. 
Our work is also related to some methods in domain adaptation and multi-lingual correlation, such as that of \citet{Bollegala:14}.

Mathematically, this work builds upon the linear algebraic understanding of modern word embeddings developed by \citet{Arora:18b} via an extension to the latent-variable embedding model of \citet{Arora:16}.
Although there have been several other applications of this model for natural language representation \cite{Arora:17,Mu:18}, ours is the first to provide a general approach for learning semantic features using corpus context.


\section{Method Specification}\label{sec:specification}

We begin by assuming a large text corpus $\mathcal{C}_\mathcal{V}$ consisting of contexts $c$ of words $w$ in a vocabulary $\mathcal{V}$, with the contexts themselves being sequences of words in $\mathcal{V}$ (e.g. a fixed-size window around the word or feature).
We further assume that we have trained word embeddings ${\bf v}_w\in\mathbb{R}^d$ on this collocation information using a standard algorithm (e.g. word2vec / GloVe).
Our goal is to construct a good embedding ${\bf v}_f\in\mathbb{R}^d$ of a text feature $f$ given a set $\mathcal{C}_f$ of contexts it occurs in.
Both $f$ and its contexts are assumed to arise via the same process that generates the large corpus $\mathcal{C}_\mathcal{V}$.
In many settings below, the number $|\mathcal{C}_f|$ of contexts available for a feature $f$ of interest is much smaller than the number $|\mathcal{C}_w|$ of contexts that the typical word $w\in\mathcal{V}$ occurs in.
This could be because the feature is rare (e.g. unseen words, $n$-grams) or due to limited human annotation (e.g. word senses, named entities).

\subsection{A Linear Approach}\label{subsec:approach}

A naive first approach to construct feature embeddings using context is {\em additive}, i.e. taking the average over all contexts of a feature $f$ of the average word vector in each context:
\begin{equation}\label{eq:additive}
{\bf v}_f^\textrm{additive}=\frac{1}{|\mathcal{C}_f|}\sum\limits_{c\in\mathcal{C}_f}\frac{1}{|c|}\sum\limits_{w\in c}{\bf v}_w
\end{equation}
This formulation reflects the training of commonly used embeddings, which employs additive composition to represent the context \cite{Mikolov:13,Pennington:14}. 
It has proved successful in the bag-of-embeddings approach to sentence representation \cite{Wieting:16,Arora:17}, which can compete with LSTM representations, and has also been given theoretical justification as the {\em maximum a posteriori} (MAP) context vector under a generative model related to popular embedding objectives \cite{Arora:16}.
\citet{Lazaridou:17} use this approach to learn embeddings of unknown word amalgamations, or {\em chimeras}, given a few context examples.

\begin{figure}[t!]
	\includegraphics[scale=0.51]{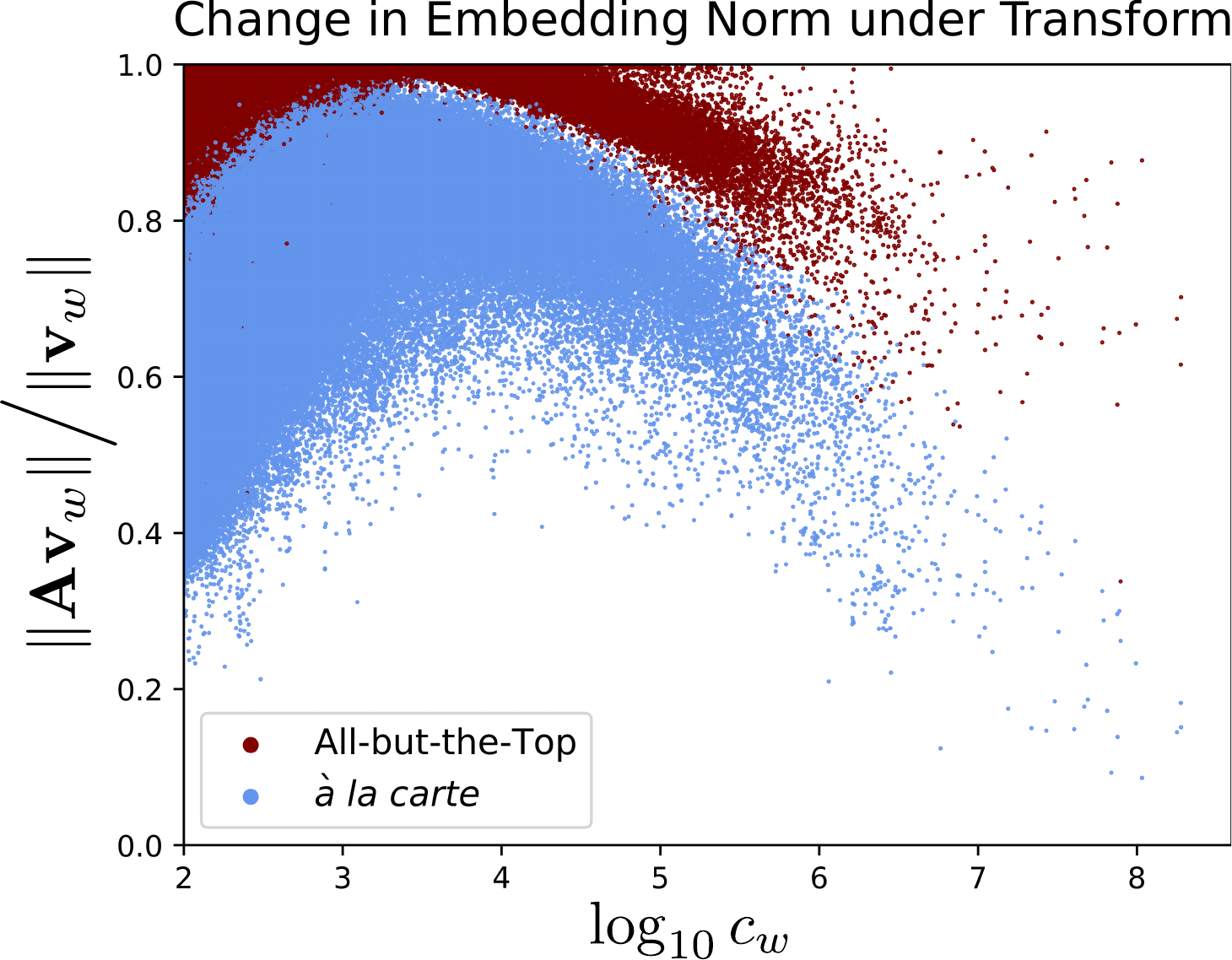}
	\caption{\label{fig:avov}
		Plot of the ratio of embedding norms after transformation as a function of word count.
		While All-but-the-Top tends to affect only very frequent words, {\em\`a la carte} learns to remove components even from less common words.
	}
\end{figure}

The additive approach has some limitations because the set of all word vectors is seen to share a few common directions. 
Simple addition amplifies the component in these directions, at the expense of less common directions that presumably carry more ``signal."
Stop-word removal can help to ameliorate this \cite{Lazaridou:17,Herbelot:17}, but does not deal with the fact that content-words also have significant components in the same direction as these deleted words.
Another mathematical framework to address this lacuna is to remove the top one or top few principal components, either from the word embeddings themselves \cite{Mu:18} or from their summations \cite{Arora:17}.
However, this approach is liable to either not remove enough noise or cause too much information loss without careful tuning (c.f. Figure~\ref{fig:avov}).

We now note that removing the component along the top few principal directions is tantamount to multiplying the additive composition by a fixed (but data-dependent) matrix.
Thus a natural extension is to use an arbitrary linear transformation which will be {\em learned} from the data, and hence guaranteed to do at least as well as any of the above ideas.
Specifically, we find the transform that can best recover {\em existing} word vectors ${\bf v}_w$ ---which are presumed to be of high quality--- from their additive context embeddings ${\bf v}_w^\textrm{additive}$.
This can be posed as the following linear regression problem
\begin{equation}\label{eq:regression}
	{\bf v}_w\approx {\bf Av}_w^\textrm{additive}={\bf A}\left(\frac{1}{|\mathcal{C}_w|}\sum\limits_{c\in\mathcal{C}_w}\sum\limits_{w'\in c}{\bf v}_{w'}\right)
\end{equation}
where ${\bf A}\in\mathbb{R}^{d\times d}$ is learned and we assume for simplicity that $\frac{1}{|c|}$ is constant (e.g. if $c$ has a fixed window size) and is thus subsumed by the transform.
After learning the matrix, we can embed any text feature in the same semantic space as the word embeddings via the following expression:
\begin{equation}\label{eq:alacarte}
	{\bf v}_f={\bf Av}_f^\textrm{additive}={\bf A}\left(\frac{1}{|\mathcal{C}_f|}\sum\limits_{c\in\mathcal{C}_f}\sum\limits_{w\in c}{\bf v}_w\right)
\end{equation}
Note that ${\bf A}$ is fixed for a given corpus and set of pretrained word embeddings and so does not need to be re-computed to embed different features or feature types. 

\begin{algorithm*}[!t]
	\DontPrintSemicolon
	\KwData{
		vocabulary $\mathcal{V}$, corpus $\mathcal{C}_\mathcal{V}$, vectors ${\bf v}_w\in\mathbb{R}^d~~\forall~w\in\mathcal{V}$, feature $f$, corpus $\mathcal{C}_f$ of contexts of $f$}
	\KwResult{feature embedding ${\bf v}_f\in\mathbb{R}^d$}
	\For{$w\in\mathcal{V}$}{
		let $\mathcal{C}_w\subset\mathcal{C}_\mathcal{V}$ be the subcorpus of contexts of $w$\\
		${\bf u}_w\gets\frac{1}{|\mathcal{C}_w|}\sum\limits_{c\in\mathcal{C}_w}\sum\limits_{w'\in c}{\bf v}_{w'}$ \tcp*{compute each word's context embedding ${\bf u}_w$}
	}
	${\bf A}\gets\argmin\limits_{{\bf A}\in\mathbb{R}^{d\times d}}\sum\limits_{w\in\mathcal{V}}\|{\bf v}_w-{\bf Au}_w\|_2^2$ \tcp*{compute context-to-feature transform ${\bf A}$}
	${\bf u}_f\gets\frac{1}{|\mathcal{C}_f|}\sum\limits_{c\in\mathcal{C}_f}\sum\limits_{w\in c}{\bf v}_w$ \tcp*{compute feature's context embedding ${\bf u}_f$}
	${\bf v}_f\gets{\bf Au}_f$ \tcp*{transform feature's context embedding}
	\caption{\label{alg:induction}
		The basic {\em\`a la carte} feature embedding induction method. All contexts $c$ consist of sequences of words drawn from the vocabulary $\mathcal{V}$.
	}
\end{algorithm*}

\paragraph{Theoretical Justification:}
As shown by \citet[Theorem~1]{Arora:18b}, the approximation \eqref{eq:regression} holds exactly in expectation for some matrix ${\bf A}$ when contexts $c\in\mathcal{C}$ are generated by sampling a context vector ${\bf v}_c\in\mathbb{R}^d$ from a zero-mean Gaussian with fixed covariance and drawing $|c|$ words using $\mathbb{P}(w|{\bf v}_c)\propto\exp\langle{\bf v}_c,{\bf v}_w\rangle$.
The correctness (again in expectation) of \eqref{eq:alacarte} under this model is a direct extension.
\citet{Arora:18b} use large text corpora to verify their model assumptions, providing theoretical justification for our approach. 
We observe that the best linear transform ${\bf A}$ can recover vectors with mean cosine similarity as high as 0.9 or more with the embeddings used to learn it, thus also justifying the method empirically.

\subsection{Practical Details}\label{subsec:details}

The basic {\em\`a la carte} method, as motivated in Section~\ref{subsec:approach} and specified in Algorithm~\ref{alg:induction}, is straightforward and parameter-free (the dimension $d$ is assumed to have been chosen beforehand, along with the other parameters of the original word embeddings).
In practice we may wish to modify the regression step in an attempt to learn a better transformation matrix ${\bf A}$.
However, the standard first approach of using $\ell_2$-regularized (Ridge) regression instead of simple linear regression gives little benefit, even when we have more parameters than word embeddings (i.e. when $d^2>|\mathcal{V}|$).

A more useful modification is to weight each point by some non-decreasing function $\alpha$ of each word's corpus count $c_w$, i.e. to solve
\begin{equation}\label{eq:weighted}
	{\bf A}=\argmin\limits_{{\bf A}\in\mathbb{R}^{d\times d}}\sum\limits_{w\in\mathcal{V}}\alpha(c_w)\|{\bf v}_w-{\bf Au}_w\|_2^2
\end{equation}
where ${\bf u}_w$ is the additive context embedding.
This reflects the fact that more frequent words likely have better pretrained embeddings.
In settings where $|\mathcal{V}|$ is large we find that a hard threshold ($\alpha(c)={\bf 1}_{c\ge\tau}$ for some $\tau\ge1$) is often useful.
When we do not have many embeddings we can still give more importance to words with better embeddings via a function such as $\alpha(c)=\log c$, which we use in Section~\ref{subsec:wsd}.


\section{One-Shot and Few-Shot Learning of Word Embeddings}\label{sec:fewshot}

While we can use our method to embed any type of text feature, its simplicity and effectiveness is rooted in word-level semantics: the approach assumes pre-existing high quality word embeddings and only considers collocations of features with words rather than with other features.
Thus to verify that our approach is reasonable we first check how it performs on word representation tasks, specifically those where word embeddings need to be learned from very few examples.
In this section we first investigate how representation quality varies with number of occurrences, as measured by performance on a similarity task that we introduce.
We then apply the {\em\`a la carte} method to two tasks measuring the ability to learn new or synthetic words from context, achieving strong results on the nonce task of \citet{Herbelot:17}.

\subsection{Similarity Correlation vs. Sample Size}\label{subsec:wordsim}

Performance on pairwise word similarity tasks is a standard way to evaluate word embeddings, with success measured via the Spearman correlation between a human score and the cosine similarity between word vectors.
An overview of widely used datasets is given by \citet{Faruqui:14}.
However, none of these datasets can be used directly to measure the effect of word frequency on embedding quality, which would help us understand the data requirements of our approach.
We address this issue by introducing the {\em Contextual Rare Words} (CRW) dataset, a subset of 562 pairs from the Rare Word (RW) dataset \cite{Luong:13} supplemented by 255 sentences (contexts) for each rare word sampled from the Westbury Wikipedia Corpus (WWC) \cite{Shaoul:10}.
In addition we provide a subset of the WWC from which all sentences containing these rare words have been removed.
The task is to use embeddings trained on this subcorpus to induce rare word embeddings from the sampled contexts.

More specifically, the CRW dataset is constructed using all pairs from the RW dataset where the rarer word occurs between 512 and 10000 times in WWC; this yields a set of 455 distinct rare words.
The lower bound ensures that we have a sufficient number of rare word contexts, while the upper bound ensures that a significant fraction of the sentences from the original WWC remain in the subcorpus we provide.
In CRW, the first word in every pair is the more frequent word and occurs in the subcorpus, while the second word occurs in the 255 sampled contexts but not in the subcorpus.
We provide word2vec embeddings trained on all words occurring at least 100 times in the WWC subcorpus;
these vectors include those assigned to the first (non-rare) words in the evaluation pairs.

\paragraph{Evaluation:}
For every rare word the method under consideration is given eight disjoint subsets containing $1,2,4,\dots,128$ example contexts.
The method induces an embedding of the rare word for each subset, letting us track how the quality of rare word vectors changes with more examples.
We report the Spearman $\rho$ (as described above) at each sample size, averaged over 100 trials obtained by shuffling each rare word's 255 contexts.

\begin{figure}[t!]
	\centering
	\includegraphics[scale=0.51]{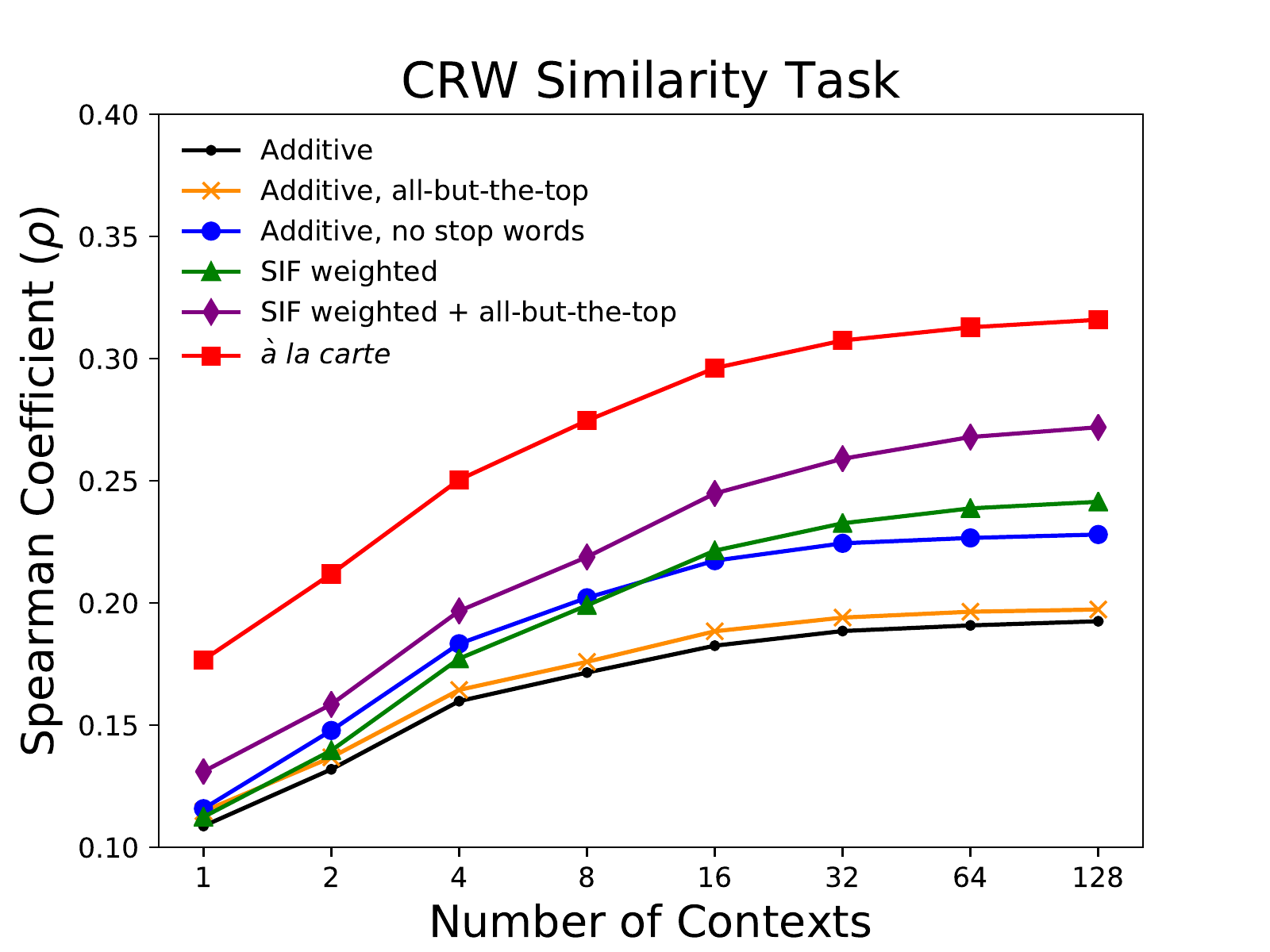}
	\caption{\label{fig:crwplot} Spearman correlation between cosine similarity and human scores for pairs of words in the CRW dataset given an increasing number of contexts per rare word. 
	Our {\em\`a la carte} method outperforms all previous approaches, even when restricted to only eight example contexts.
	}
\end{figure}

The results in Figure~\ref{fig:crwplot} show that our {\em\` a la carte} method significantly outperforms the additive baseline \eqref{eq:additive} and its variants, including stop-word removal, SIF-weighting \cite{Arora:17}, and top principal component removal \cite{Mu:18}.
We find that combining SIF-weighting and top component removal also beats these baselines, but still does worse than our method.
These experiments consolidate our intuitions from Section~\ref{sec:specification} that removing common components and frequent words is important and that learning a data-dependent transformation is an effective way to do this.
However, if we train word2vec embeddings from scratch on the subcorpus together with the sampled contexts we achieve a Spearman correlation of 0.45;
this gap between word2vec and our method shows that there remains room for even better approaches for few-shot learning of word embeddings.

\subsection{Learning Embeddings of New Concepts: Nonces and Chimeras}\label{subsec:nonce}

\begin{table*}[t!]
	\centering
	\begin{threeparttable}
		\begin{tabular}{lcccccc}
			& \multicolumn{2}{c}{Nonce \cite{Herbelot:17}} && \multicolumn{3}{c}{Chimera \cite{Lazaridou:17}}\\
			Method & ~Mean Recip. Rank~ & Med. Rank && ~~2 Sent.~~ & ~~4 Sent.~~ & 6 Sent. \\
			\toprule
			word2vec & 0.00007 & 111012 && 0.1459 & 0.2457 & 0.2498 \\
			additive & 0.00945 & 3381 && 0.3627 & 0.3701 & 0.3595 \\
			additive, no stop words & 0.03686 & 861 && 0.3376 & 0.3624 & \bf0.4080 \\
			nonce2vec & 0.04907 & 623 && 0.3320 & 0.3668 & 0.3890 \\
			{\em\`a la carte} & \bf0.07058 & \bf165.5 && \bf0.3634 & \bf0.3844 & 0.3941 \\
			\bottomrule
		\end{tabular}
	\end{threeparttable}
	\caption{\label{tbl:nonce}
		Comparison with baselines and nonce2vec \cite{Herbelot:17} on few-shot embedding tasks.
		Performance on the chimeras task is measured using the Spearman correlation with human ratings. Note that the additive baseline requires removing stop-words in order to improve with more data.
	}
\end{table*}

We now evaluate our work directly on the tasks posed by \citet{Herbelot:17}, who developed simple datasets and methods to ``simulate the process by which a competent speaker encounters a new word in known contexts."
The general goal will be to construct embeddings of new concepts in the same semantic space as a known embedding vocabulary using contextual information consisting of definitions or example sentences.

\paragraph{Nonces:}
We first discuss the definitional nonce dataset made by the authors themselves, which has a test-set consisting of 300 single-word concepts and their definitions.
The task of learning each concept's embedding is simulated by removing or randomly re-initializing its vector and requiring the system to use the remaining embeddings and the definition to make a new vector that is close to the original.
Because the embeddings were constructed using data that includes these concepts, an implicit assumption is made that including or excluding one word does not greatly affect the semantic space;
this assumption is necessary in order to have a good target vector for the system to be evaluated against.

Using 259,376 word2vec embeddings trained on Wikipedia as the base vectors, \citet{Herbelot:17} heavily modify the skip-gram algorithm to successfully learn on one definition, creating the {\em nonce2vec} system.
The original skip-gram algorithm and ${\bf v}_w^\textrm{additive}$ are used as baselines, with performance measured as the mean reciprocal rank and median rank of the concept's original vector among the nearest neighbors of the output.

To compare directly to their approach, we use their word2vec embeddings along with contexts from the Wikipedia corpus to construct context vectors ${\bf u}_w$ for all words $w$ apart from the 300 nonces.
We then learn the {\em\`a la carte} transform ${\bf A}$, weighting the data points in the regression \eqref{eq:weighted} using a hard threshold of at least 1000 occurrences in Wikipedia. 
An embedding for each nonce can then be constructed by multiplying ${\bf A}$ by the sum over all word embeddings in the nonce's definition.
As can be seen in Table~\ref{tbl:nonce}, this approach significantly improves over both baselines and nonce2vec; 
the median rank of 165.5 of the original embedding among the nearest neighbors of the nonce vector is very low considering the vocabulary size is more than 250,000, and is also significantly lower than that of all previous methods.

\paragraph{Chimeras:} 
The second dataset \citet{Herbelot:17} consider is that of \citet{Lazaridou:17}, who construct unseen concepts by combining two related words into a fake nonce word (the ``chimera") and provide two, four, or six example sentences for this nonce drawn from sentences containing one of the two component words. 
The desired nonce embeddings is then evaluated via the correlation of its cosine similarity with the embeddings of several other words, with ratings provided by human judges.

We use the same approach as in the nonce task, except that the chimera embedding is the result of summing over multiple sentences.
From Table~\ref{tbl:nonce} we see that, while our method is consistently better than both the additive baseline and nonce2vec, removing stop-words from the additive baseline leads to stronger performance for more sentences.
Since the {\em\`a la carte} algorithm explicitly trains the transform to match the true word embedding rather than human similarity measures, it is perhaps not surprising that our approach is much more dominant on the definitional nonce task.


\section{Building Feature Embeddings using Large Corpora}

Having witnessed its success at representing unseen words, we now apply the {\em\`a la carte} method to two types of feature embeddings: synset embeddings and $n$-gram embeddings.
Using these two examples we demonstrate the flexibility and adaptability of our approach when handling different corpora, base word embeddings, and downstream applications.

\subsection{Supervised Synset Embeddings for Word-Sense Disambiguation}\label{subsec:wsd}

\begin{table*}[t!]
	\centering
	\begin{threeparttable}
		\begin{tabular}{lcccccccc}
			&& SemEval-2013 Task 12 && \multicolumn{5}{c}{SemEval-2015 Task 13}\\
			Method && nouns && adj. & nouns & adv. & verbs & comb. \\
			\toprule
			{\em\`a la carte} (SemCor) && 60.0 && 72.2 & 67.7 & 85.2 & 60.6 & 68.1 \\
			{\em\`a la carte} (glosses) && 51.8 && 75.3 & 62.5 & 79.0 & 55.8 & 64.2 \\
			{\em\`a la carte} (combined) && \bf60.5 && 74.1 & \bf70.3 & 86.4 & 59.4 & \bf69.6\\
			\midrule
			MFS (SemCor) && 58.8 && \bf79.5 & 60.0 & \bf87.6 & \bf 66.7 & 66.8 \\
			\citet{Raganato:17} && \underline{\bf66.9} && & & & & \underline{\bf72.4} \\
			\bottomrule
		\end{tabular}
	\end{threeparttable}
	\caption{\label{tbl:wsd}
		Application of {\em\`a la carte} synset embeddings to two standard WSD tasks.
		As all systems always return exactly one answer, performance is measured in terms of accuracy.
		Results due to \citet{Raganato:17}, who use a bi-LSTM for this task, are given as the recent state-of-the-art result.
	}
\end{table*}

Embeddings of synsets, or sets of cognitive synonyms, and related entities such as senses and lexemes have been widely studied, often due to the desire to account for polysemy \cite{Rothe:15,Iacobacci:15}.
Such representations can be evaluated in several ways, including via their use for word-sense disambiguation (WSD), the task of determining a word's sense from context.
While current state-of-the-art methods often use powerful recurrent models \cite{Raganato:17}, we will instead use a simple similarity-based approach that heavily depends on the synset embedding itself and thus serves as a more useful indicator of representation quality.
A major target for our simple systems is to beat the most-frequent sense (MFS) method, which returns for each word the sense that occurs most frequently in a corpus such as SemCor.
This baseline is ``notoriously hard-to-beat," routinely besting many systems in SemEval WSD competitions \cite{Navigli:13}.

\paragraph{Synset Embeddings:}

We use SemCor \cite{Langone:04}, a subset of the Brown Corpus (BC) \cite{Francis:79} annotated using PWN synsets.
However, because the corpus is quite small we use GloVe trained on Wikipedia instead of on BC itself.
The transform ${\bf A}$ is learned using context embeddings ${\bf u}_w$ computed with windows of size ten around occurrences of $w$ in BC and weighting each word by the log of its count during the regression stage \eqref{eq:weighted}.
Then we set the context embedding ${\bf u}_s$ of each synset $s$ to be the average sum of word embeddings representation over all sentences in SemCor containing $s$.
Finally, we apply the {\em\` a la carte} transform to get the synset embedding ${\bf v}_s={\bf Au}_s$.

\paragraph{Sense Disambiguation:}
To determine the sense of a word $w$ given its context $c$, we convert $c$ into a vector using the {\em\`a la carte} transform ${\bf A}$ on the sum of its word embeddings and return the synset $s$ of $w$ whose embedding ${\bf v}_s$ is most similar to this vector.
We try two different synset embeddings: those induced from SemCor as above and those obtained by embedding a synset using its {\em gloss}, or PWN-provided definition, in the same way as a nonce in Section~\ref{subsec:nonce}.
We also consider a {\em combined} approach in which we fall back on the gloss vector if the synset does not appear in SemCor and thus has no induced embedding.

As shown in Table~\ref{tbl:wsd}, synset embeddings induced from SemCor alone beat MFS overall, largely due to good noun results.
The method improves further when combined with the gloss approach.
While we do not match the state-of-the-art, our success in besting a difficult baseline using very little fine-tuning and exploiting none of the underlying graph structure suggests that the {\em\`a la carte} method can learn useful synset embeddings, even from relatively small data.

\subsection{N-Gram Embeddings for Classification}

\begin{table*}[t!]
	\centering
	\begin{threeparttable}
		\begin{tabular}{lccccc}
			Method && beef up & cutting edge & harry potter & tight lipped \\
			\toprule
			${\bf v}_{w_1}+{\bf v}_{w_2}$ && meat, out & cut, edges & deathly, azkaban & loose, fitting \\
			${\bf v}_{(w_1,w_2)}^\textrm{additive}$ && but, however & which, both & which, but & but, however \\
			ECO && meats, meat & weft, edges & robards, keach & scaly, bristly \\
			Sent2Vec && add, reallocate & science, multidisciplinary & naruto, pokemon & wintel, codebase \\
			{\em\`a la carte} && need, improve & innovative, technology & deathly, hallows & worried, very \\
			\bottomrule
		\end{tabular}
	\end{threeparttable}
	\caption{\label{tbl:bigrams}
		Closest word embeddings (measured via cosine similarity) to the embeddings of four idiomatic or entity-associated bigrams.
		From these examples we see that purely compositional methods may struggle to construct context-aware bigram embeddings, even when the features are present in the corpus.
		On the other hand, adding up corpus contexts \eqref{eq:additive} is dominated by stop-word information.
		Sent2Vec is successful on half the examples, reflecting its focus on good sentence, not bigram, embeddings.
	}
\end{table*}
 
As some of the simplest and most useful linguistic features, $n$-grams have long been a focus of embedding studies.
Compositional approaches, such as sums and products of unigram vectors, are often used and work well on some evaluations, but are often order-insensitive or very high-dimensional \cite{Mitchell:10}.
Recent work by \citet{Poliak:17} works around this while staying compositional;
however, as we will see their approach does not seem to capture a bigram's meaning much better than the sum of its word vectors.
$n$-grams embeddings have also gained interest for low-dimensional document representation schemes \cite{Hill:16,Pagliardini:18,Arora:18a}, largely due to the success of their sparse high-dimensional Bag-of-$n$-Grams (BonG) counterparts \cite{Wang:12}.
This setting of document embeddings derived from $n$-gram features will be used for quantitative evaluation in this section.

We build $n$-gram embeddings using two corpora: 300-dimensional Wikipedia embeddings, which we evaluate qualitatively, and 1600-dimensional embeddings on the Amazon Product Corpus \cite{McAuley:15}, which we use for document classification.
For both we use as source embeddings GloVe vectors trained on the respective corpora over words occurring at least a hundred times.
Context embeddings are constructed using a window of size ten and a hard threshold at 1000 occurrences is used as the word-weighting function in the regression \eqref{eq:weighted}.
Unlike \citet{Poliak:17}, who can construct arbitrary embeddings but need to train at least two sets of vectors of dimension at least $2d$ to do so, and \citet{Yin:14}, who determine which $n$-grams to represent via corpus counts, our {\em\`a la carte} approach allows us to train exactly those embeddings that we need for downstream tasks.
This, combined with our method's efficiency, allows us to construct more than two million bigram embeddings and more than five million trigram embeddings, constrained only by their presence in the large source corpus.

\begin{table*}[t!]
	\centering
	\begin{threeparttable}
		\small
		\begin{tabular}{@{}lcccccccccccc@{}}
			Representation && $n$ & $d^\ast$ && MR & CR & SUBJ & MPQA & TREC & SST $(\pm1)$ & SST & IMDB \\
			\toprule
			
			\multirow{3}*{BonG}
			&& 1 & $V_1$ && 77.1 & 77.0 & 91.0 & 85.1 & 86.8 & 80.7 & 36.8 & 88.3 \\
			&& 2 & $V_1+V_2$ && 77.8 & 78.1 & 91.8 & 85.8 & 90.0 & 80.9 & 39.0 & 90.0 \\
			&& 3 & $V_1+V_2+V_3$ && 77.8 & 78.3 & 91.4 & 85.6 & 89.8 & 80.1 & 42.3 & 89.8 \\
			\midrule
			
			\multirow{3}*{\em\`a la carte}
			&& 1 & 1600 && 79.8 & 81.3 & 92.6 & 87.4 & 85.6 & 84.1 & 46.7 & 89.0 \\
			&& 2 & 3200 && 81.3 & 83.7 & 93.5 & 87.6 & 89.0 & 85.8 & \bf47.8 & \bf90.3 \\
			&& 3 & 4800 && \bf81.8 & \bf84.3 & 93.8 & 87.6 & 89.0 & \bf86.7 & \underline{\bf48.1} & \underline{\bf90.9} \\
			\midrule
			
			$\textrm{Sent2Vec}^1$ && 1-2 & 700 && 76.3 & 79.1 & 91.2 & 87.2 & 85.8 & 80.2 & 31.0 & 85.5 \\
			$\textrm{DisC}^2$ && 2-3 & 3200-4800 && 80.1 & 81.5 & 92.6 & 87.9 & 90.0 & 85.5 & 46.7 & 89.6 \\
			\midrule
			
			$\textrm{skip-thoughts}^3$ &&& 4800 && 80.3 & 83.8 & \bf94.2 & \bf88.9 & \underline{\bf93.0} & 85.1 & 45.8 \\
			$\textrm{SDAE}^4$ &&& 2400 && 74.6 & 78.0 & 90.8 & 86.9 & 78.4 \\
			$\textrm{CNN-LSTM}^5$ &&& 4800 && 77.8 & 82.0 & 93.6 & \bf89.4 & \bf92.6 & \\
			$\textrm{MC-QT}^6$ &&& 4800 && \underline{\bf82.4} & \underline{\bf86.0} & \underline{\bf94.8} & \underline{\bf90.2} & \bf92.4 & \underline{\bf87.6} \\
			\midrule
			
			$\textrm{byte mLSTM}^7$ &&& 4096 && \bf86.8 & \bf90.6 & \bf94.7 & 88.8 & 90.4 & \bf91.7 & \bf54.6 & \bf92.2 \\
			
			\bottomrule
		\end{tabular}
		\begin{tablenotes}
			\item[$\ast$] Vocabulary sizes (i.e. BonG dimensions) vary by task; usually 10K-100K.
			\item[1,3,7]\cite{Pagliardini:18,Kiros:15,Radford:17} Evaluation conducted using latest pretrained models. Note that the latest available skip-thoughts implementation returns an error on the IMDB task.
			\item[2,4,5,6]\cite{Arora:18a,Hill:16,Gan:17,Logeswaran:18} Best results from publication.
		\end{tablenotes}
	\end{threeparttable}
	\caption{\label{tbl:perf}
		Performance of document embeddings built using {\em\`a la carte} $n$-gram vectors and recent unsupervised word-level approaches on classification tasks, with the character LSTM of \cite{Radford:17} shown for comparison.
		Top three results are {\bf bolded} and the best word-level performance is \underline{underlined}.
	}
\end{table*}

\paragraph{Qualitative Evaluation:}
We first compare bigram embedding methods by picking some idiomatic and entity-related bigrams and examining the closest word vectors to their representations.
These word-pairs are picked because we expect sophisticated feature embedding methods to encode a better vector than the sum of the two embeddings, which we use as a baseline.
From Table~\ref{tbl:bigrams} we see that embeddings based on corpora rather than composition are better able to embed these bigrams to be close to concepts that are semantically similar.
On the other hand, as discussed in Section~\ref{sec:specification} and evident from these results, the additive context approach is liable to emphasize stop-word directions due to their high frequency.

\paragraph{Document Embedding:}
Our main application and quantitative evaluation of $n$-gram vectors is to use them to construct document embeddings.
Given a length $L$ document $D=\{w_1,\dots,w_L\}$, we define its embedding ${\bf v}_D$ as a weighted concatenation over sums of our induced $n$-gram embeddings, i.e.
\begin{equation*}
{\bf v}_D^T=\begin{pmatrix}\sum\limits_{t=1}^L{\bf v}_{w_t}^T&\cdots&\frac{1}{n}\sum\limits_{t=1}^{L-n+1}{\bf v}_{(w_t,\dots,w_{t+n-1})}^T\end{pmatrix}
\end{equation*}
where ${\bf v}_{(w_t,\dots,w_{t+n-1})}$ is the embedding of the $n$-gram $(w_t,\dots,w_{t+n-1})$.
Following \citet{Arora:18a}, we weight each $n$-gram component by $\frac{1}{n}$ to reflect the fact that higher-order $n$-grams have lower quality embeddings because they occur less often in the source corpus.
While we concatenate across unigram, bigram, and trigram embeddings to construct our text representations, separate experiments show that simply adding up the vectors of all features also yields a smaller but still substantial improvement over the unigram performance.
The higher embedding dimension due to concatenation is in line with previous methods and can also be theoretically supported as yielding a less lossy compression of the $n$-gram information \cite{Arora:18a}.

In Table~\ref{tbl:perf} we display the result of running cross-validated, $\ell_2$-regularized logistic regression on documents from MR movie reviews \citep{Pang:05}, CR customer reviews \citep{Hu:04}, SUBJ subjectivity dataset \citep{Pang:04}, MPQA opinion polarity subtask \citep{Wiebe:05}, TREC question classification \citep{Li:02}, SST sentiment classification (binary and fine-grained) \citep{Socher:13}, and IMDB movie reviews \citep{Maas:11}.
The first four are evaluated using tenfold cross-validation, while the others have train-test splits.

Despite the simplicity of our embeddings (a concatenation over sums of {\em\` a la carte} $n$-gram vectors), we find that our results are very competitive with many recent unsupervised methods, achieving the best word-level results on two of the tested datasets.
The fact that we do especially well on the sentiment tasks indicates strong exploitation of the Amazon review corpus, which was also used by DisC, CNN-LSTM, and byte mLSTM.
At the same time, the fact that our results are comparable to neural approaches indicates that local word-order may contain much of the information needed to do well on these tasks.
On the other hand, separate experiments do not show a substantial improvement from our approach over unigram methods such as SIF \cite{Arora:17} on sentence similarity tasks such as STS \cite{Cer:17}.
This could reflect either noise in the $n$-gram embeddings themselves or the comparative lower importance of local word-order for textual similarity compared to classification.

\section{Conclusion}
We have introduced {\em\`a la carte} embedding, a simple method for representing semantic features using unsupervised context information.
A natural and principled integration of recent ideas for composing word vectors, the approach achieves strong performance on several tasks and promises to be useful in many linguistic settings and to yield many further research directions.
Of particular interest is the replacement of simple window contexts by other structures, such as dependency parses, that could yield results in domains such as question answering or semantic role labeling.
Extensions of the mathematical formulation, such as the use of word weighting when building context vectors as in \citet{Arora:18b} or of spectral information along the lines of \citet{Mu:18}, are also worthy of further study.

More practically, the Contextual Rare Words (CRW) dataset we provide will support research on few-shot learning of word embeddings.
Both in this area and for $n$-grams there is great scope for combining our approach with compositional approaches \cite{Bojanowski:16,Poliak:17} that can handle settings such as zero-shot learning.
More work is needed to understand the usefulness of our method for representing (potentially cross-lingual) entities such as synsets, whose embeddings have found use in enhancing WordNet and related knowledge bases \cite{Camacho:16,Khodak:17}.
Finally, there remain many language features, such as named entities and morphological forms, whose representation by our method remains unexplored.

\section*{Acknowledgments}
We thank Karthik Narasimhan and our three anonymous reviewers for helpful suggestions.
The work in this paper was in part supported by SRC JUMP, Mozilla Research, NSF grants CCF-1302518 and CCF-1527371, Simons Investigator Award, Simons Collaboration Grant, and ONR-N00014-16-1-2329.

\bibliography{reference}
\bibliographystyle{acl_natbib}

\end{document}